\documentclass[nohyperref]{article}

\usepackage{microtype}
\usepackage{graphicx}
\usepackage{subfigure}
\usepackage{booktabs} 

\usepackage{hyperref}

\usepackage[final]{mytemplate}

\usepackage{amsmath}
\usepackage{amssymb}
\usepackage{mathtools}
\usepackage{amsthm}
\usepackage{bm}

\usepackage[capitalize,noabbrev]{cleveref}

\theoremstyle{plain}

\theoremstyle{definition}

\theoremstyle{remark}

\usepackage[textsize=tiny]{todonotes}

\title{Convolutional Neural Processes for Inpainting Satellite Images}

\author{Alexander Pondaven$^*$
  \\\bfseries Hamzah Hashim$^\dagger$
  \And M\"{a}rt Bakler$^*$
  \\\bfseries Martin Ignatov
  \And Donghu Guo$^\dagger$
  \\\bfseries Harrison Zhu
  \AND
  \normalfont
  Imperial College London \\
  \texttt{\{ap2619,mb1221,dg321,hh2019,mgi18,hbz15\}@ic.ac.uk}
}

\begin{document}
\maketitle

{
\renewcommand\thefootnote{}\footnotetext[0]{\textsuperscript{\ensuremath{*}}Equal contribution. $\dagger$ \text{Order decided via a coin toss}.}
}

\begin{abstract} 
    The widespread availability of satellite images has allowed researchers to model complex systems such as disease dynamics. However, many satellite images have missing values due to measurement defects, which render them unusable without data imputation. For example, the scanline corrector for the LANDSAT 7 satellite broke down in 2003, resulting in a loss of around 20\% of its data. Inpainting involves predicting what is missing based on the known pixels and is an old problem in image processing, classically based on PDEs or interpolation methods, but recent deep learning approaches have shown promise. However, many of these methods do not explicitly take into account the inherent spatiotemporal structure of satellite images. In this work, we cast satellite image inpainting as a natural meta-learning problem, and propose using convolutional neural processes (ConvNPs) where we frame each satellite image as its own task or 2D regression problem.  We show ConvNPs can outperform classical methods and state-of-the-art deep learning inpainting models on a scanline inpainting problem for LANDSAT 7 satellite images, assessed on a variety of in and out-of-distribution images.
    
\end{abstract}

\section{Introduction}
\begin{figure}[t]
\centering
\vspace{-2.8mm}
\subfigure[Before 31st of May 2003]{
\includegraphics[width=0.47\textwidth]{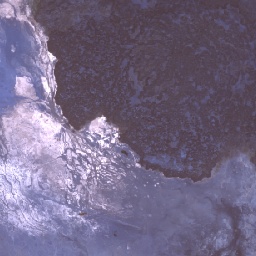}
}
\subfigure[After 31st of May 2003]{
\includegraphics[width=0.47\columnwidth]{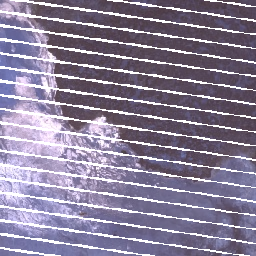}
}
\caption{LANDSAT 7 images before and after the scanline corrector failure.}
\label{fig:raw_img}
\end{figure}

Satellite images are a valuable resource to research communities. With the surge of computational methods using remote sensing data, satellite images have been widely used for instance in epidemiology \citep{weiss2019mapping}, social sciences \citep{yeh2020using} and crop yield modelling \citep{zhu2022}. One of the satellites that is commonly used is the LANDSAT 7 \citep{landsat7} satellite due to its long temporal coverage and high spatial resolution. However, due to a mechanical fault in the satellite's scanline corrector (SLC), satellite imagery post 31st May 2003 suffers from lines of missing and corrupted pixels (Figure \ref{fig:raw_img}). As they occupy a significant area of the satellite images (about 20\% of the data), the images obtained from LANDSAT 7 lost much of their research use as the scanlines significantly impair the performance of computational methods using the images.

To `repair' the images, inpainting techniques must be used. Image inpainting, also known as gap-filling, aims to fill the corrupted pixels in an image with values that resemble the original pixel values as closely as possible. Inpainting for corrupted image data is an active area of research and many methods have been introduced. For instance, many deterministic methods have been proposed, that use higher order differential equations \citep{, burger2009cahn, bertalmio2001navier, bertozzi2011unconditionally}. Moreover, recent advances in deep learning have shown promising results in image inpainting. \citet{ronneberger2015u} introduced the celebrated U-Net, which was originally used for biomedical image segmentation, but can be trained to perform image inpainting, and \citet{partialconv} introduced Partial Convolutions (PartialConv), which is a modification to the classical convolutional layer to make it suitable for inpainting.

One drawback of the traditional deep learning methods is that they treat all of the images as a single task. They approximate a single function $f_\theta$ that is learned by the network and it is used during inference to get the predictive values $\tilde{y} = f_\theta(x)$. This approach does not take into account the spatiotemporal differences between different images, where different predictive functions $f_{\theta_m}$ could better suit different tasks $m$. This kind of problem may be better suited to meta-learning methods, which learn task-specific representations and are better at capturing the differences between various inputs. 

In meta-learning, we learn a task representation that can determine the exact function for each distinct task, $\tilde{y} = f_{\theta_m}(x)$, where $m$ is the current task. \citet{garnelo2018conditional} introduced a meta-learning approach called Conditional Neural Processes (CNPs) that uses an encoder-decoder architecture to obtain a predictive distribution $p_{\theta_m}(y \vert x)$, which is equivalent to obtaining a distribution over predictive functions $f_{\theta_m}$. \citet{gordon2019convolutional} and \citet{foong2020meta} introduced Convolutional Conditional Neural Processes (ConvCNPs) and Convolutional Latent Neural Processes (ConvLNPs) respectively, which are better suited for image inpainting tasks due to their translational equivariance property. These Convolutional Neural Processes (ConvNPs)  are shown to exhibit very good few-shot and zero-shot learning capabilities as well, which has been demonstrated for inpainting weather data \citep{foong2020meta, markou2022practical}.

In this paper, we show that ConvNPs can be used for satellite image inpainting, in particular, to correct the scanlines of LANDSAT 7 images. We use ConvCNPs and ConvLNPs with an MS-SSIM similarity score loss function \citep{wang2004image} that is better for generating sharp images. We show that our ConvNPs can outperform  many state-of-the-art image inpainting models and accurate inpainting models can be trained with a relatively small dataset (training set of 800 images with dimensions 128x128 or 64x64). ConvNP models also show good performance for both in-distribution (inpainting of similar images as they are trained on) and out-of-distribution (OOD) satellite images (zero-shot tasks, images of different regions than the model was trained on). The consequence of the latter is that we are then able to construct a \textbf{global inpainter} despite only training with images from a \textbf{small subset} of spatiotemporal locations. In addition, via a downstream synthetic CNN regression problem, with inputs being the imputed images, we show that using the ConvLNP imputed images yield the closest results to regression over the original clean images.

\begin{figure}[t]
\centering
  \includegraphics[width=0.4\textwidth,height=6cm]{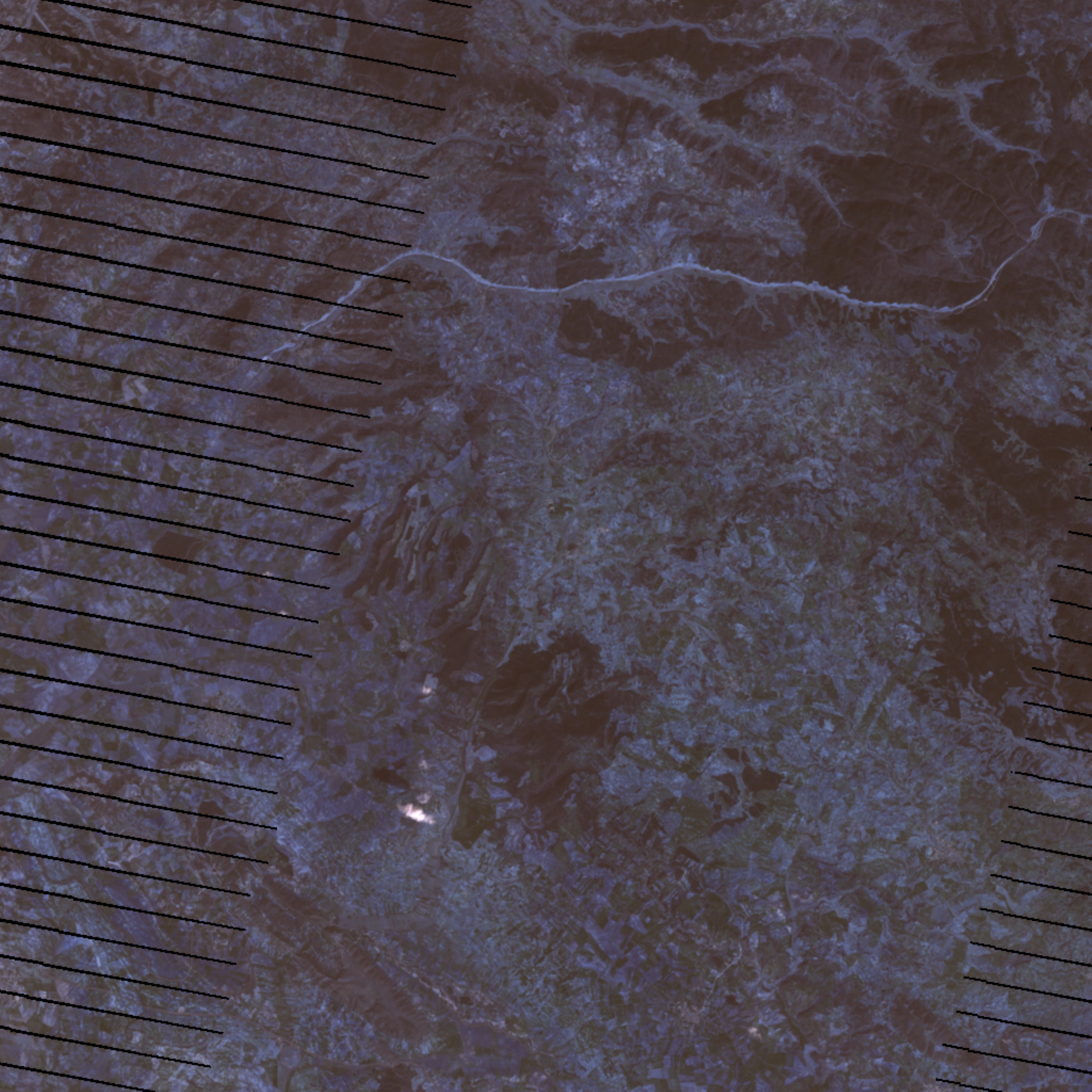}
  \includegraphics[width=0.4\textwidth,height=6cm]{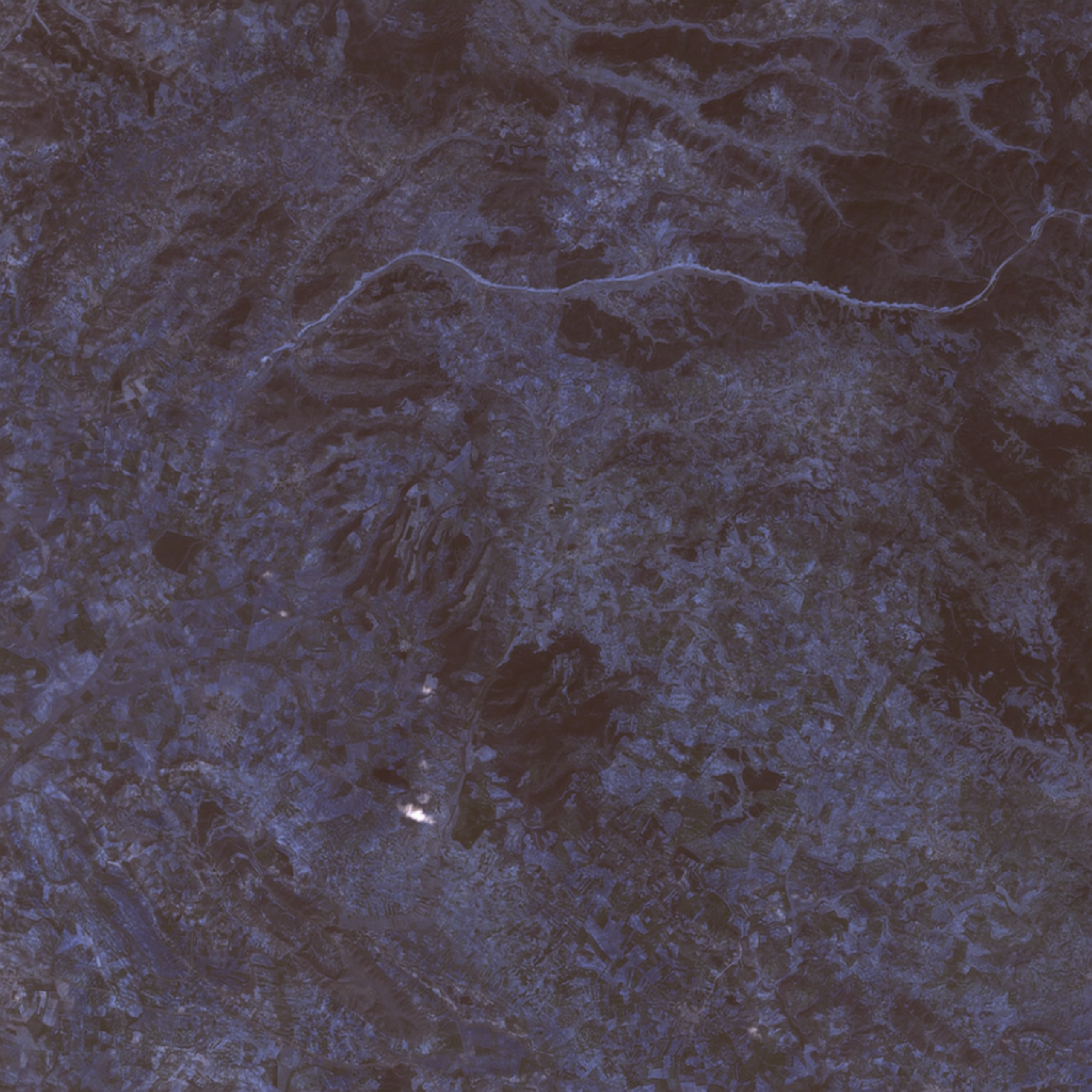}
  \caption{Kenya 1024x1024 by predicting on 64x64 patches with ConvCNP. (Left) Original image (Right) Inpainted image.}
  \label{patches}
\end{figure}

\section{Related work}
\paragraph{Classical approaches:}
Single source methods involve inpainting based on the undamaged pixels in the image. Traditional methods include interpolation and PDE or convolution based methods like Bertalmio's, Telea's and Oliveira's algorithms \citep{bertalmio2001navier, telea2004image, richard2001fast}, but these methods only inpaint by looking at neighbouring pixels within each image. Exemplar-based inpainting matches patches in a specific order \citep{exemplar_survey}, but these methods use defined low-level patterns that do not capture semantic structure in the images. The Navier-Stokes \citep{bertalmio2001navier} inpainting algorithm uses ideas from classical fluid dynamics and is used as a baseline for our problem. It is deterministic and does not require training, but the method does not utilise or learn any prior knowledge of the data distribution. In addition, \citet{scaramuzza2005landsat} introduced an inpainting algorithm specifically for LANDSAT 7, which compared pre- and post-2003 images for scanline filling in multiple phases, and is adopted for use as one of the official LANDSAT 7 products. However, the data-engineering required for this method is extremely complicated and we are unable to reproduce it.

\paragraph{Deep learning approaches:}
Deep learning approaches to inpainting are more globally consistent and achieve better local details.  Many of these take the form of encoder-decoder CNNs, such as U-Net \citep{ronneberger2015u} or GAN-based algorithms \cite{ContextEncoder}. U-Net consists of a U-shaped CNN that first downsamples the input image using convolutions and pooling layers, and then upsamples using transposed convolutions and skip-connections. Neural Path Synthesis \cite{MSNPS} enhances the GAN-based approach with a texture network inspired by style transfer to transfer the style of the context pixels to improve local details. Multi-scale discriminators have been employed as well to capture global and local details \cite{GLCIC,PatchGAN}. 

Another approach is to mask the missing pixels and enhance the network with custom operations on the masked pixels.
Inpainting irregular holes has seen success using this approach by using Partial Convolutions \cite{partialconv}, where the proposed architecture introduces novel partial convolution operations and blocks with custom loss functions, which are specially designed to take into account missing data. VAE networks like HI-VAEs \citep{nazabal2020handling} have been shown to be suitable for problems with missing data and can infer masked values. However, these models require large datasets and long training regimes, which make them difficult to generalise to new spatiotemporal locations.

Recent developments such as \citet{Dupont2021GenerativeMA,dupont2022} use a continuous representation for images and pixels by modelling them as functions rather than discrete values, which can also be used for imputing missing values.  \citet{lugmayr2022repaint} introduced RePaint, which uses denoising diffusion probabilistic models and produces high-quality inpainted images. Finally, \citet{foong2020meta} and \citet{markou2022practical} both perform image inpainting for Earth observation data, specifically gridded weather data, using ConvNPs. This problem is very similar to inpainting satellite images, although our aim is to inpaint scanlines after the SLC failure of the LANDSAT 7 satellite and we propose to use a more suitable likelihood/loss function.

\section{Data}
LANDSAT 7 images are downloaded using the Google Earth Engine (GEE) API \cite{gorelick2017google}. For this paper, we focus solely on the visible RGB bands of the LANDSAT 7 satellite (B3, B2, B1) with spatial resolutions of 30 meters. The images are sampled from a uniform spatial grid with a grid spacing of 0.4 degrees longitude and latitude. They are downloaded with a dimension of 256x256 pixels, corresponding to a land area of approximately $59 \text{km}^2$. Due to computational limitations, these images are cropped to 64x64 and 128x128, and model results are reported using these sizes. Note that smaller satellite images could be patched together to perform a larger inpainting task as seen in Figure~\ref{patches} and Appendix~\ref{appendix}. Satellite images from Kenya are used for training. Data from UK, Norway, Brazil and Nepal are also collected to test the model's capabilities on out-of-distribution (unseen, location-wise) images.

The images are extracted from specific dates and locations, and are divided into pre- and post-2003 (when the SLC broke). All pre-2003 data is from between 1999 to 2003. Post-2003 data is collected from between 2003 to 2004. Images with missing pixels (alpha channel is present) are filtered out for pre-2003 images. UK images are sometimes completely white due to the presence of clouds, which resulted in `better' inpainting results across all models. To create more challenging out-of-distribution tasks, Norway, Brazil and Nepal images are filtered by only taking images where the middle 64x64 section had less than 90\% white pixels (so the cropped 64x64 dimension dataset also had less clouds). Post-2003 data is just used to extract a set of 100 scanline bit masks from Kenya data to apply to un-corrupted images during training. Some images had large sections of missing pixels, so the post-2003 images are filtered to have $< 20\%$ missing pixels, but also at least 100 missing pixels (set arbitrarily) to ensure the presence of scanlines.

\section{Methodology}
We cast satellite inpainting as a meta-learning problem. The pixel locations on the grid and RGB pixel values at those locations are denoted as $x\in\mathbb{R}^2$ and $y\in\mathbb{R}^3$ respectively. We denote the context set of pixels as $(x_C,y_C):=\{x_{i}, y_{i} \}_{i=1}^{N_C}$, and the target set as $(x_T,y_T):=\{\bar{x}_{i}, \bar{y}_{i} \}_{i=1}^{N_T}$. The union of the context and target set represents the task $D:=\{C, T\}$, where $C=\{x_C, y_C\}$ and $T=\{x_T, y_T\}$. Each task $D_m$ corresponds to an image, which could also be viewed as a 2D function \citep{Dupont2021GenerativeMA, dupont2022}. At prediction time, $x_C,x_T$ and $y_C$ are observed but $y_T$ is not.

Classical inpainting methods would assume that we learn a global function $f_\theta$ that gives a prediction $y_T\approx f_\theta(x_T)$. Similarly, U-Net and PartialConv would not explicitly, but \textbf{implicitly}, distinguish between different tasks and require enormous training sets with data augmentation in order to learn network weights such that $f_\theta(x_{C_m},y_{C_m},x_{T_m})\approx f_{\theta_m}(x_{T_m})$. We argue that taking the meta-learning viewpoint allows us to \textbf{explicitly} take into account the spatiotemporal variations for each task and thus promote efficient learning.

Meta-learning methods \citep{thrun2012learning, finn2017model} aim to solve the problem of using a distinct function at inference time to predict target set values. In our setting, during training, we learn a global parameter $\theta$, which, given contexts of a task, could also output a \textbf{task-specific} representation $R_m$. The global objective function is given by $\mathbb{E}_{m\sim \mathcal{M}}[\mathcal{L}(D_{\eta}(E_{\xi}(x_{C_m}, y_{C_m}))(x_{T_m}), y_T)]$, with $ D_{\eta}(E_{\xi}(x_{C_m}, y_{C_m}))(x_{T_m})\approx f_{\theta_m}(x_{T_m})$, where $\theta=(\eta,\xi)$, $E_\xi$ encodes the context set $(x_C,y_C)$ to a task-specific representation, $D_\eta$ decodes the task-specific representation and target location to the output, and $\mathcal{L}$ is a loss function.

This results in a model that adjusts the predictor function $f$ depending on the context set of the task. One advantage of meta learning is that the predictive functions use both the information from the current context set of the task as well as the information that is shared across tasks, making the method well-suited for OOD tasks. This allows modelling of heterogeneous function distributions and is a beneficial property for satellite image inpainting as they have multiple zero-shot tasks for different spatial locations and times, that are not seen during training.

\paragraph{Neural Processes:} Conditional \citep{garnelo2018conditional} and latent neural processes \citep{np}, as a wider family of Neural Processes (NPs), employ a general encoder-decoder neural network architecture that enables meta-learning of functions or stochastic processes. The encoder $E_\xi$ takes as input the context points $C$ and outputs a task representation $R=E_\xi(C)$, which gets passed to the decoder $D_\eta(R, \cdot)$ to give a task-specific output function distribution. This makes it suitable for satellite image inpainting since the context points $C$ are the pixels that are not missing, and the remaining target points $T$ are pixels covered by the scanlines.

\paragraph{Convolutional Conditional Neural Processes:}
However, a shortfall of standard Conditional NPs  \citep{garnelo2018conditional} for image inpainting is the lack of translation equivariance, which is an important property to have for image data. \citet{gordon2019convolutional} introduced translational equivariance to the NP family through ConvCNPs. For our purposes, we are solely interested in the ConvCNPs for on-the-grid data (e.g.\ images). With the same notation, we denote the original image as $I$ and the context mask $M_C$, for which $[M_C]_{i,j}=1$ if the pixel at location $(i,j)$ is in the context set, and 0 otherwise. Our masked context set is thus given by $Z_C=M_C \odot I$. Concatenating the context mask and the masked context point, we thus get $\phi = [M_C, Z_C]^\intercal$. Applying a convolution to $\phi$, we obtain the functional representation $R = \text{Conv}_\theta([M_C, Z_C]^\intercal)$, where $\text{Conv}_\theta$ is the 2D convolution operator with positively-constrained kernel parameters $\theta$. We then apply the normalisation $R^{(1:C)} = R^{(1:C)} / R^{0}$. This step is known as \textbf{SetConv} (when not evaluated at the target points). We can decode $R$ using a CNN, which includes an absorbed MLP to map the output of the CNN at each location $(i,j)$ to $\mathbb{R}^2$ and gives $\bm{\mu}$, the image prediction.

\begin{figure}[t]
\centering
\includegraphics[width=0.4\textwidth]{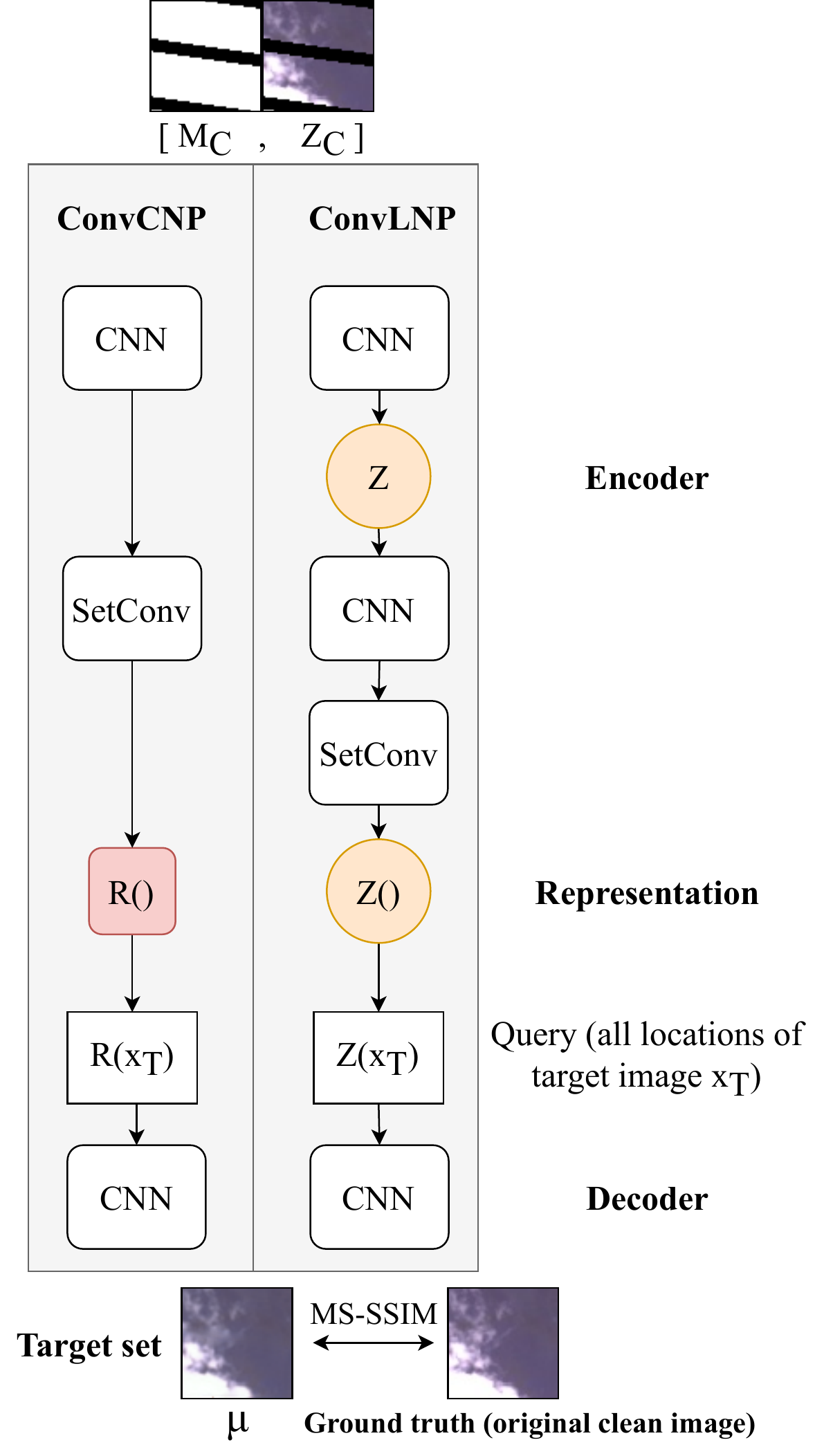}
  \includegraphics[width=0.45\textwidth]{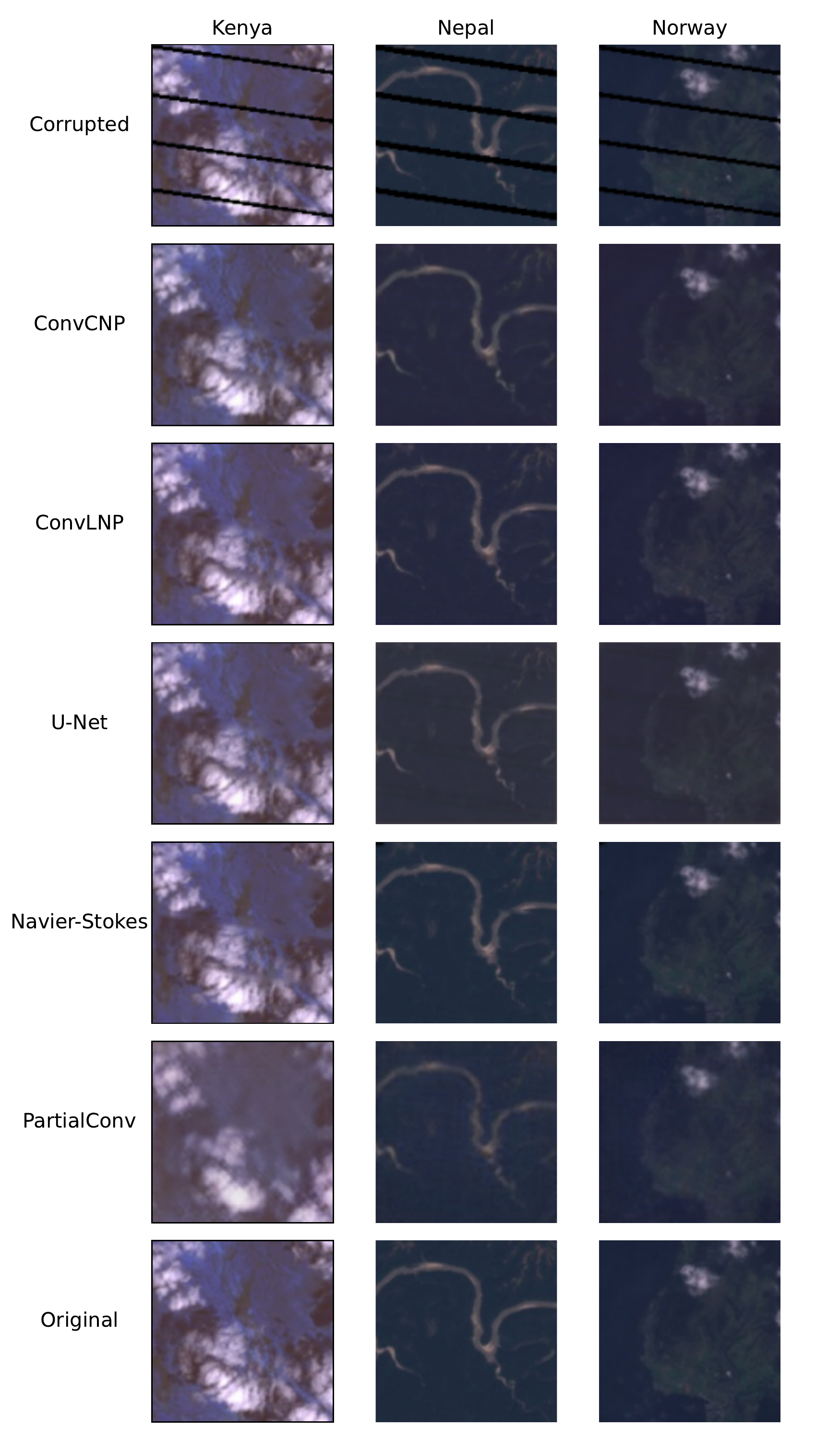}
  \caption{(Left) ConvCNP and ConvLNP on-the-grid architecture. (Right) Inpainting predictions for all models on 128x128 images (all region predictions in Figure~\ref{predictions}).}
  \vspace{-3mm}
  \label{predictons_mini}
\end{figure}

\paragraph{Convolutional Latent Neural Processes:}
\citet{foong2020meta} presents the Convolutional Neural Processes (ConvNPs, in this paper referred to as Convolutional Latent Neural process or ConvLNP) that utilise a latent variable to capture information from the context set. It is similar in architecture to the conditional neural process with the encoder-decoder architecture, but in the ConvLNP the encoder outputs a distribution over the latent variable \textbf{z} with the SetConv representations: $\mathbf{z} \sim p(\mathbf{z} \vert R)$. This enables ConvCNPs to learn `richer joint predictive distributions' \citep{foong2020meta} and handle multimodalities. ConvLNP can be straighforwardly implemented on top of ConvCNPs by using the SetConv representations to parameterise the Gaussian latent variable distribution, and then decode the latent variable samples using a CNN. The full computational graphs for ConvCNP and ConvLNP are described in Figure~\ref{predictons_mini}.

\paragraph{Training objective:}

Following \citet{foong2020meta}, we use the maximum likelihood training objective for the ConvNPs: $\mathcal{L} = \log p(y_T \vert x_T, C)$, which measures the predictive performance of the models. For the log-likelihood function, $\log p(y_T \vert x_T, C)$, we instead use the MS-SSIM metric (Multi Scale Structural Similarity, \citet{wang2003multiscale}) between the mean predictions and ground truth images. MS-SSIM is a structural similarity metric for images, and is widely used in the field of signal processing, having shown empirically to increase sharpness of final prediction images. In the ConvLNP training objective, the maximum likelihood approach uses sample estimates to approximate the likelihood of the predictions: $\mathcal{L}\approx \frac{1}{L}\sum_{l=1}^L \log p(y_T \vert \mathbf{z}, x_T, C)$.

\section{Experiments}
We study the performance of ConvCNPs and ConvLNPs for the task of inpainting LANDSAT 7 scanlines. We compare the results to the baseline models consisting of Navier-Stokes (NS) inpainting algorithm, U-Net and PartialConv, for which the latter two are vastly popular and yield state-of-the-art results on a variety of image inpainting problems.  We train the ConvCNP, ConvLNP and U-Net using the MS-SSIM loss function, and for PartialConv we use the loss proposed in \citet{partialconv}. We conduct extensive experiments to measure the in-distribution performance of each model by inpainting satellite images of the same country, as well as OOD performance on a set of different countries (zero-shot prediction over unseen spatial locations). The inpainted results are also evaluated on a synthetic downstream regression task.  The extracted scanline masks are randomly applied to the datasets of pre-2003 images so that both the clean and corrupted images are available during training. For ConvNP models, we treat the pixels outside of the scanline mask as the context set. To evaluate the performance of all the models, we compute the MS-SSIM between the predictions and the ground truth images. For ConvCNP, U-Net and PartialConv, the prediction is the forward pass of the networks. For NS, the prediction is the gap-filled image. For ConvLNP, we take the mean of the model outputs over many samples from the latent variable distribution. The MS-SSIM is bounded in $[0,1]$, where values closer to 1 show that the images are more similar.

\subsection{Inpainting LANDSAT 7 Images}
\subsubsection{Data}

The training images are acquired from the LANDSAT 7 Satellite before the scanlines are present in the images. The training set consists of 1000 images from Kenya with dimensions 128x128 and 64x64. The scanlines are acquired from 100 Kenya images post-SLC failure. We perform 5-fold cross validation with a 80\%-20\% train-test ratio for each split. During training, a scanline is applied to each image as a mask chosen randomly within the 100 scanlines extracted. To test the model's performance on the in-distribution test set, each model is tested on the respective Kenya test set of that split. To test the models' zero-shot capabilities, 1000 images of UK, Nepal, Brazil and Norway are collected, for which the model had not seen during training. For each location, a clean image and a "corrupted" image is created, where the corrupted image is created by applying one of the randomly chosen extracted scanlines as a mask to a clean image.  

\smallskip

\subsubsection{ConvNP training:}
Our implementation follows \citet{dubois2020npf}. Both ConvCNPs and ConvLNPs use Resnet blocks in the encoder and linear MLPs in the decoder. The ConvCNP has a 10-layer ResNet encoder with a representation size of 128 channels and the decoder MLP has 4 layers. It is trained for 400 epochs, with batch size 8 and  learning rate $10^{-4}$, which decays exponentially by a factor of 5. The ConvLNP model for 128x128 images is trained for 200 epochs with a batch size of 4 (low batch size due to computational limitations) and during training, 4 samples are obtained of the latent variable while during evaluation, 8 samples are used. For 64x64, the latent samples are increased for ConvLNP, namely 16 latent samples are used for training and 32 are used during inference. Both Resnets used in ConvLNP have 8 layers. 

\smallskip

\subsubsection{Results}
\begin{figure}[t]
\centering
  \includegraphics[width=0.8\textwidth]{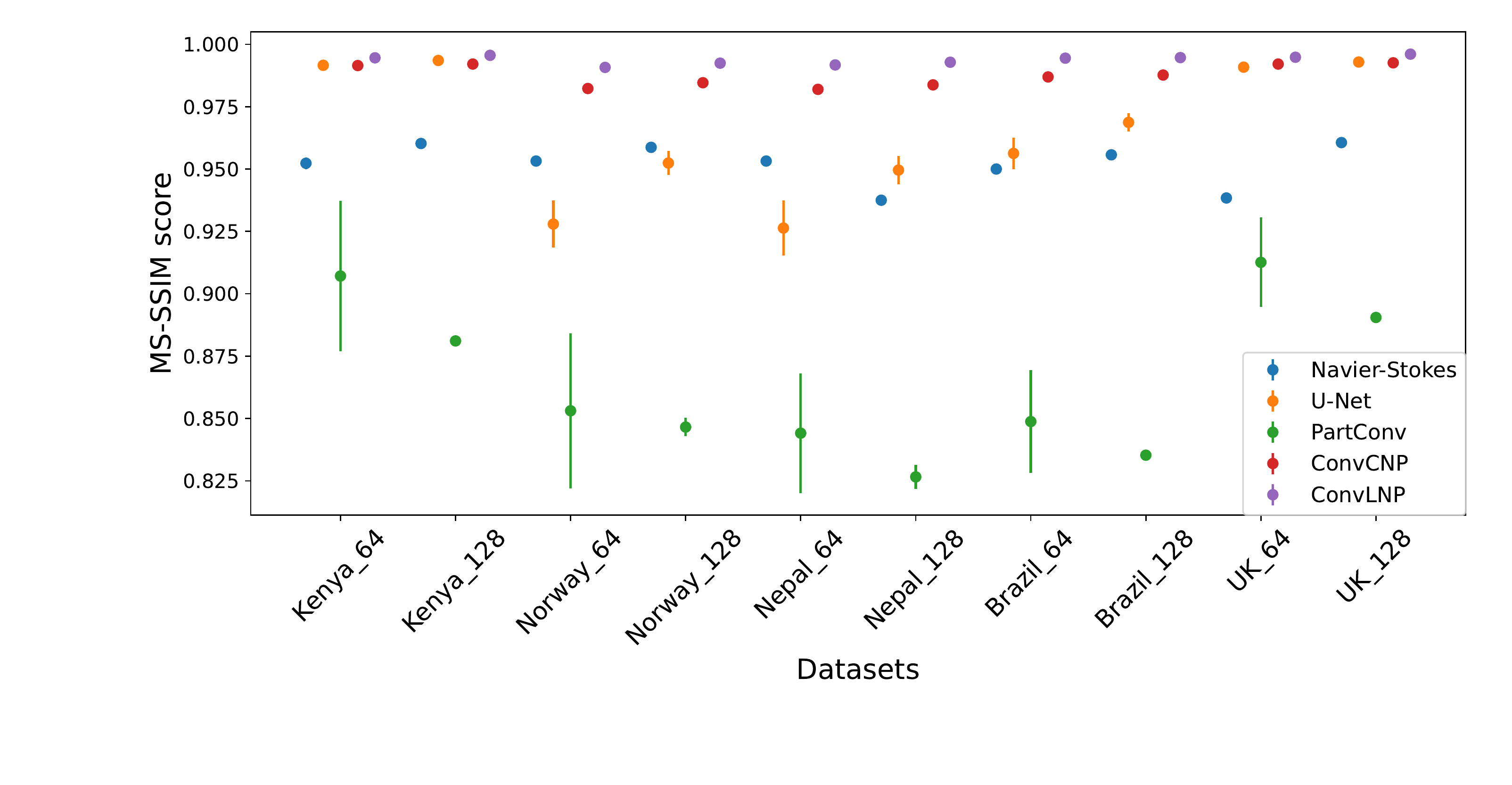}
  \vspace{-5mm}
  \caption{Mean and standard error of the MS-SSIM scores over 5-fold cross validation for predicting over Kenya and OOD datasets. Note that standard errors lower than 0.01 have not been visualised.}
  \vspace{-8mm}
  \label{res:inpainting}
\end{figure}

We first examine the MS-SSIM scores of the Neural Process models and baselines in the in-distribution setting - trained on images of Kenya and inference in a held out test-set of Kenya images. The MS-SSIM score  is calculated for 200 test images by comparing the clean image and inpainted image for all the algorithms. The results are seen in Figure~\ref{res:inpainting}, where average results across 5 splits are reported. One caveat is that we only ran PartialConv for 3 splits, due to a software issue. The HI-VAE model is also tested, but is not able to produce consistent images in our small dataset and low training time setting and hence has been omitted from the analysis. 

As can be seen both empirically and by the MS-SSIM scores in Figure~\ref{res:inpainting}, ConvNPs and U-Net achieve very good results for the Kenya test dataset. PartialConv is not able to output as good inpainting results and there is a noticeable difference between original and inpainted images in terms of image sharpness and quality of detail. The model seems to be able to successfully remove the scanlines but the resulting images are blurry compared to the original image which leads to lower MS-SSIM scores. One of the reasons could be that the model requires a longer training time; in the original paper \cite{partialconv} the model is trained for significantly longer with a significantly larger dataset. Navier-Stokes fails in particular scenarios where the target pixels are at the border between two differently coloured regions as the predicted pixel values are an average over the two regions, resulting in noticeably wrong estimations. However, ConvNPs and U-Net produce outputs with high MS-SSIM scores. The predictions are good in terms of consistency and sharpness, and scanlines are not visible.

For the second experiment,  the zero-shot or OOD capabilities of the inpainting models are examined. The models trained on the 800-image Kenya test split are evaluated on different countries including UK, Nepal, Brazil and Norway, where the model has not seen any of these locations during training. The MS-SSIM score between predictions and original images is again calculated and averaged across the 5 splits. The Navier-Stokes algorithm here cannot be considered as solving a zero-shot task as it does not involve training, but the results are shown for comparison. The average MS-SSIM scores can again be seen in Figure~\ref{res:inpainting}.

In the OOD setting, the ConvNP models outperform all the baseline models. For both U-Net and PartialConv, the MS-SSIM scores are considerably lower and scanlines are quite visible in the predictions compared to the in-distribution results. The ConvNP models, however, generalise quite well for the OOD task and predictions have high MS-SSIM scores. The scanlines are fairly well inpainted in comparison to the original pixel values. U-Net performs surprisingly well for UK images, however this is related to the high cloud presence in UK images similarly to Kenya images.
The difference in generalisation performance is due to the architecture of classical deep learning networks, which are treating all images for inpainting as a single task and are inherently approximating a single function from the input to the output. However, the image locations, times and scanline locations are different, and this approach does not model the differences between satellite images very well, which results in lower generalisation performance of the models. However, the meta-learning approach treats the input images as different tasks and hence the variability between images and their characteristics is better accounted. Therefore, we see an improvement in quality of predictions for new and unseen images.
One considerable advantage, however, of U-Net is its speed and training stability. The training is considerably faster for U-Net. For PartialConv, training speed is comparable to the ConvNP training times and the results are considerably worse, as it converges a lot slower than ConvNPs or U-Net. 

\subsection{Synthetic downstream task}
\subsubsection{Data}

\begin{figure}[t]
\centering
  \includegraphics[width=0.5\textwidth]{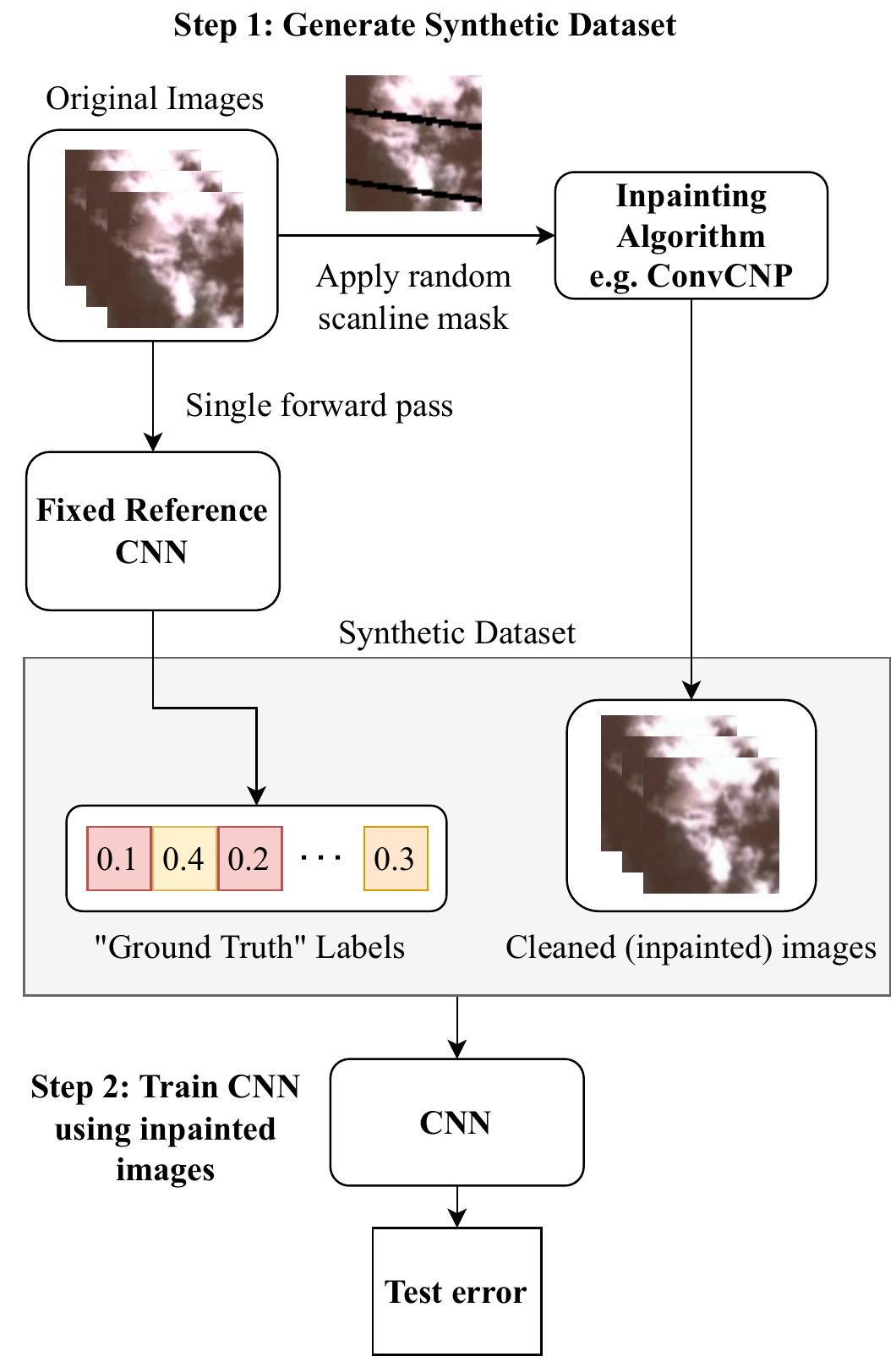}
  \caption{Synthetic downstream task workflow.}
  \vspace{-5mm}
  \label{downstream}
\end{figure}

Inpainted results from each model are evaluated on a synthetic downstream task (Figure~\ref{downstream}). A synthetic dataset is formed by passing non-corrupted images $X$ through a small randomly initialised CNN model, $f$, and scaling the images by a factor of $a=10$ to create the ground truth output for each image, $f(a*X)$. This scaling is done to make the downstream task more sensitive to different values within the scanlines. Another CNN, $g$, with the same architecture but a different random initialisation is then trained using the inpainted images $\tilde{X}$ to evaluate how well it can perform in a regression task as a substitute to the clean image. The inpainted images are created by adding a random scanline and inpainting it using the models discussed. This generates a synthetic training set: ($\tilde{X}$, $f(a*X)$). As a reference, the training performance on the original images $X$ and on the images with the scanline applied is also evaluated.

\subsubsection{Training}
All images are normalised to the range [0,1]. The CNN architecture includes two convolutional layers with kernel size 3. The output is then fed through a fully connected layer to produce a single scalar output. Training used the MSE loss and a batch size of 8. Training is done with 5-fold cross-validation and each run lasted for 300 epochs. We used a learning rate of 0.001 with a reduction of a factor of 0.1 if it plateaus with a patience of 3 epochs. Early stopping is also used with patience 8 and threshold 0.0001.

\subsubsection{Results}

\begin{figure}[t]
\centering
  \includegraphics[width=0.7\textwidth]{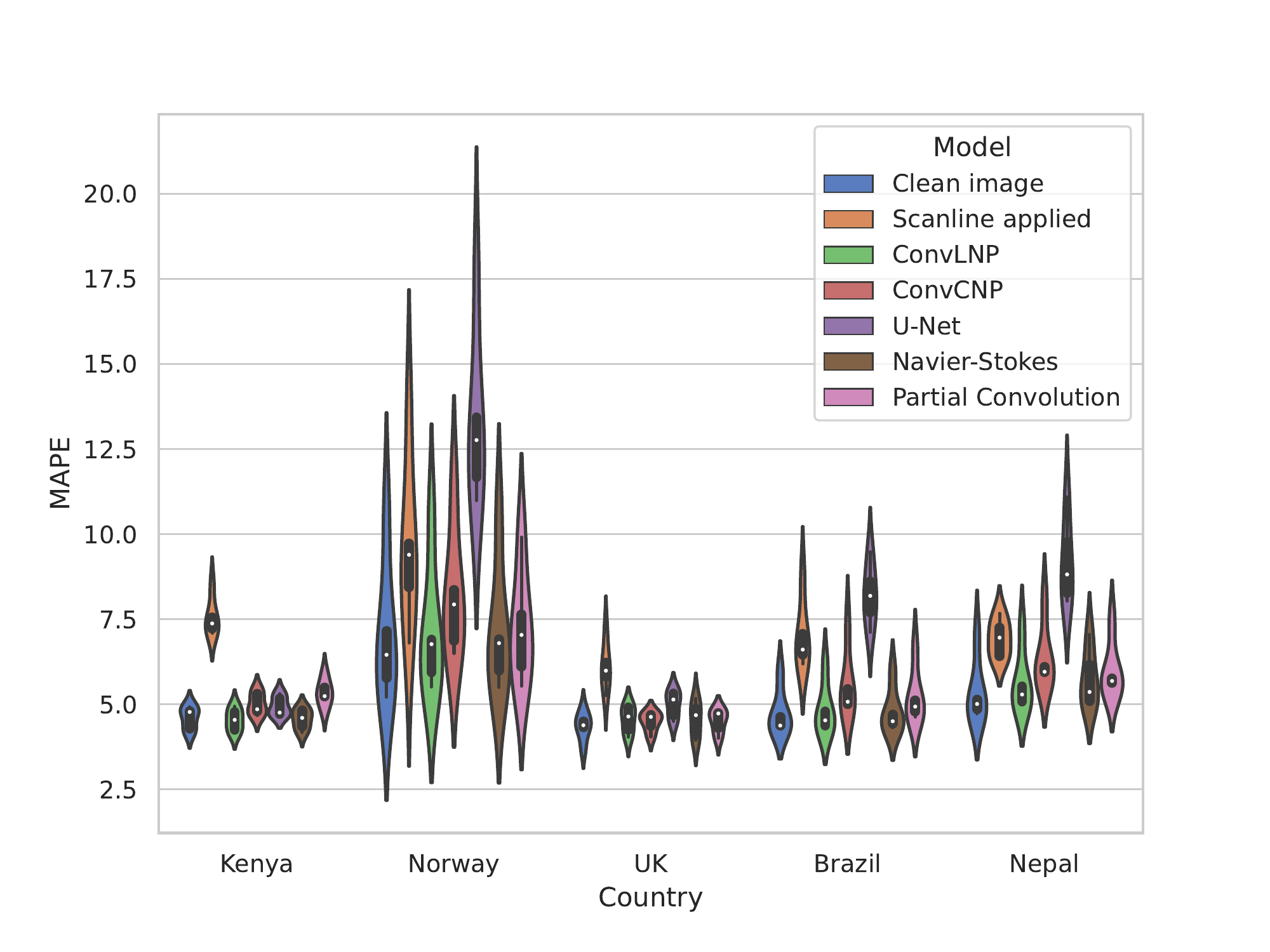}
  \caption{MAPE test errors over 5-fold cross-validation on downstream regression tasks over Kenya and OOD datasets with image dimension 64x64.}
  \label{res:downstream}
\end{figure}

The results for 64x64 satellite image predictions on the downstream task are presented in Figure~\ref{res:downstream}. The predictions for 128x128 dimensional images are unstable with the current training settings, so are omitted from the report. As expected, the clean image baseline results show the best performance as they are used to generate the synthetic dataset. For Kenya (in-distribution), ConvLNP outperform the others, performing very similarly to the clean images in the downstream task. As expected, the scanline results have the worst performance as they are not inpaint. U-Net performs similarly to ConvCNP, but for out-of-distribution tasks, U-Net performs the worst out of all models, even performing worse than running the downstream task with the scanline applied images. As U-Net only approximates a single function during inpainting, it fails at generalising to these different tasks.

Overall, ConvLNP achieves among the best MAPE errors in both in-distribution and out-of-distribution tasks. ConvCNP has slightly higher test MAPE. The ConvLNP can learn more complex data distributions than ConvCNPs by using latent variables, allowing for improved prediction results more similar to the clean image.  Navier-Stokes does get very similar MAPE, although this is likely due to most of the pixels being the exact same as the clean image. Given only the scanlines have changed, this is not an appropriate measure of the quality of inpainting predictions made using this algorithm. Surprisingly, Partial Convolution test MAPE consistently performs slightly better than ConvCNP, despite inpainting results being blurry and of worse quality. Norway satellite images seem to also be the most challenging to learn given this approach. This can be seen by the large error even with the clean image baseline.

\section{Limitations}

There are several improvements to our approach that could be done in future work. Firstly, the downstream regression task is synthetic and does not reflect the performance of imputed satellite data in real-life tasks. A more suitable approach is to evaluate predictions on an epidemiology downstream regression task such as Malaria prevalence mapping. This could involve inpainting LANDSAT 7 maps inside regions of interest to predict Malaria cases, and using models such as DeepSets \citep{deepsets}, Set Transformers \citep{set_transformer}, and distribution regression methods \citep{zhu2022}. 

Another limitation is that each original satellite image, of around $6000\times 6000$ pixels wide, has scanlines that are increasing in thickness. Our training set only uses relatively thinner scanlines after data processing, resulting in poor performance on thicker scanlines, so further exploration of performance of ConvNPs when trained on larger scanlines could be done, such as augmenting the set of training scanlines so that we include thicker ones. Finally, this work can also be readily extended to the task of cloud removal.

\section{Conclusion and Discussion}
We find that ConvNPs are successful at inpainting LANDSAT 7 satellite images corrupted by scanlines in both in-distribution and out-of-distribution tasks, outperforming classic and state-of-the-art inpainting methods. We also observe that ConvLNPs perform the best out of these models in a synthetic downstream regression task. These ConvNP models are able to take advantage of the spatiotemporal nature of satellite images to understand the underlying structure of the data. The direct consequence of this is that with ConvNPs, we may be able to obtain a \textbf{global inpainter} for LANDSAT 7, by only training on a \textbf{small subset} of spatiotemporal locations, which is computationally tractable compared to training U-Net or PartialConv over training images taken from all over Earth. Future work could involve making use of recent advances in ConvNPs to improve expressiveness \citep{bruinsma2021gaussian, markou2021efficient, markou2022practical} and to more explicitly account for space-time \citep{singh2019sequential}.

\nocite{langley00}

\section{Acknowledgements and Funding Disclosure}
Many thanks to Samir Bhatt for proposing this problem. The authors would like to thank Samir Bhatt, Seth Flaxman, Jevgenij Gamper, Iwona Hawryluk, Tom Mellan, Swapnil Mishra, Oliver Ratmann and Maxime Rischard for useful feedback and suggestions. The authors would also like to thank the Imperial College Data Science Society (ICDSS), especially Kenton Kwok, for organising and supporting this project as part of the Advanced Team. HZ is supported by the EPSRC Centre for Doctoral Training in Modern Statistics and Statistical Machine Learning (EP/S023151/1), the Department of Mathematics of Imperial College London and Cervest Limited. We would also like to thank Andy Thomas for his endless support with using the NVIDIA4 GPU Compute Server. 

\bibliographystyle{icml2022}
\bibliography{ref}

\newpage
\appendix

\section{Additional Experimental Results}
\label{appendix}

\begin{figure}[hbtp]
\centering
  \includegraphics[width=\textwidth]{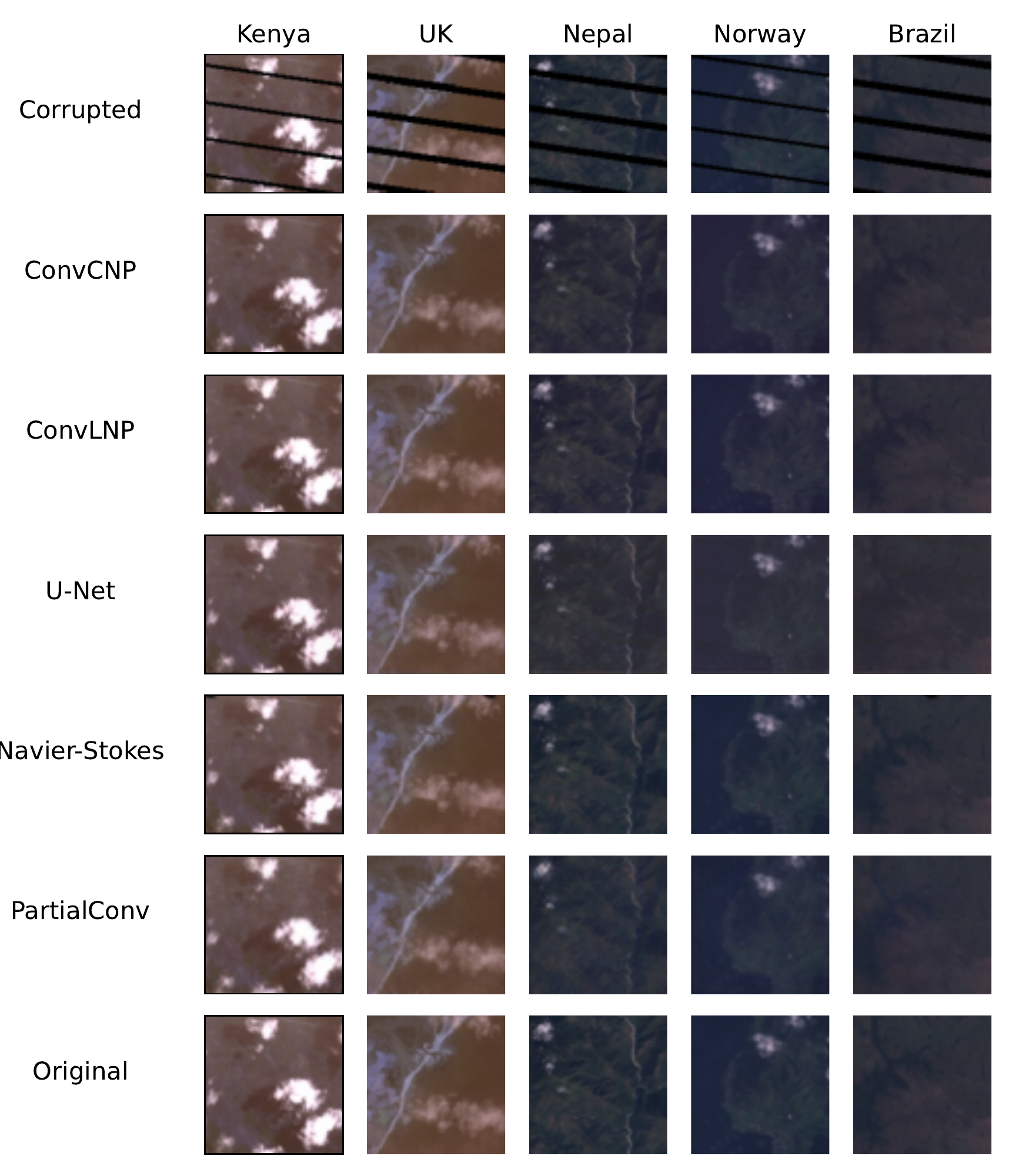}
  \caption{Inpainting results on 128x128 images for all models over multiple regions.}
  \label{predictions}
\end{figure}

\begin{figure}[hbtp]
\centering
  \includegraphics[width=\textwidth]{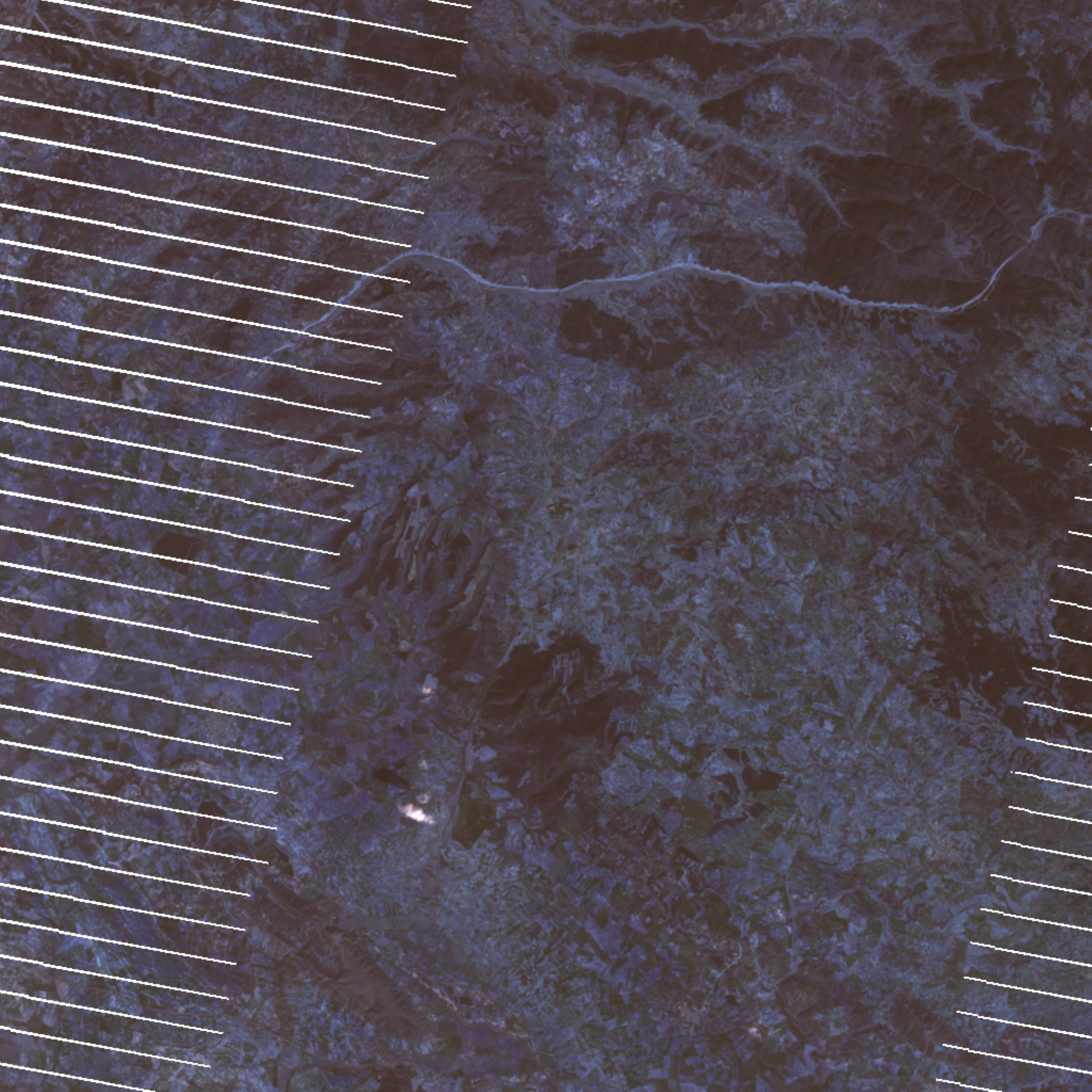}
  \caption{Corrupted 1024x1024 Kenya image used for inpainting of figures in this section.}
  \label{patches_corrupted}
\end{figure}

\begin{figure}[hbtp]
\centering
  \includegraphics[width=\textwidth]{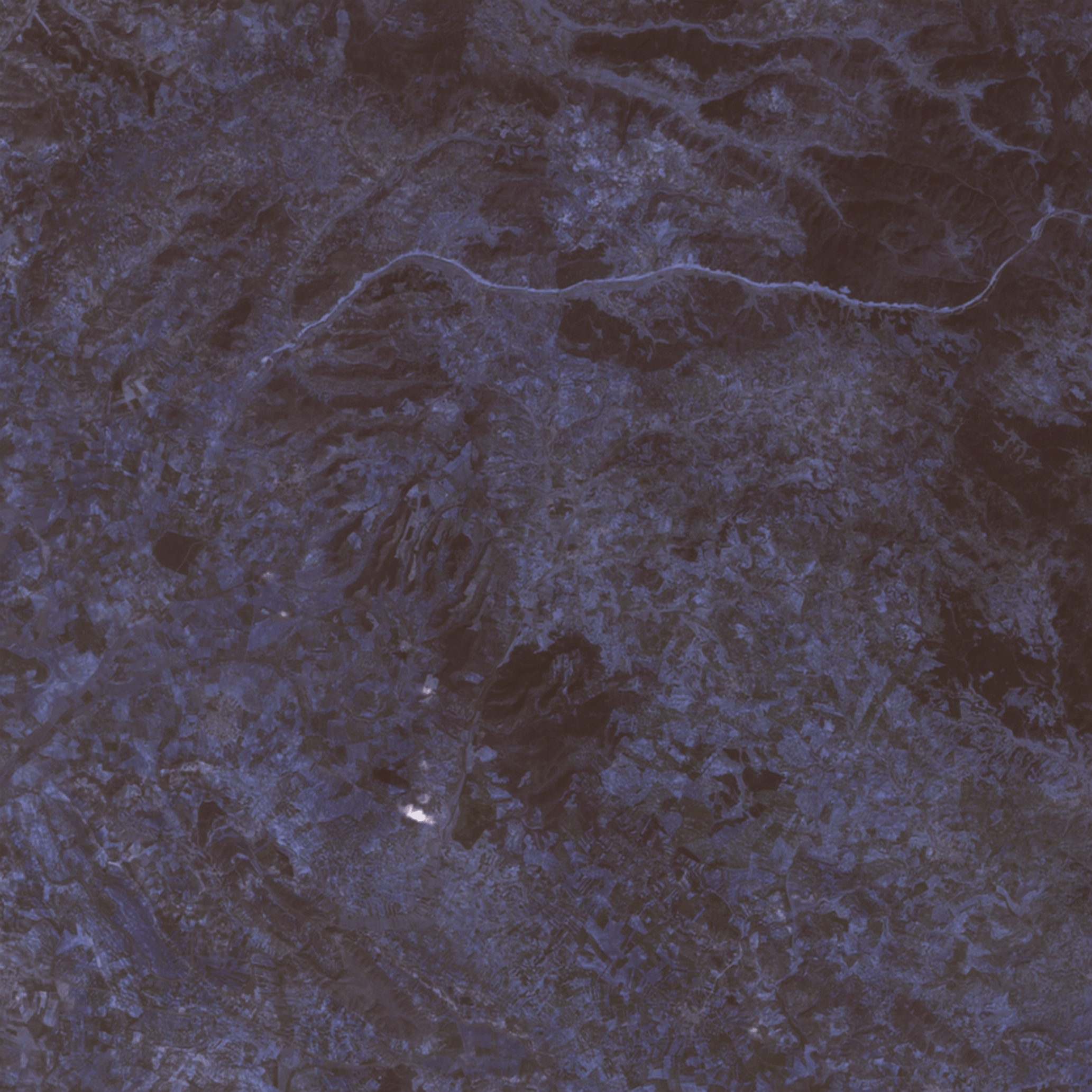}
  \caption{ConvCNP inpainting of 1024x1024 Kenya image using 64x64 patches.}
  \label{patches_convcnp}
\end{figure}

\begin{figure}[hbtp]
\centering
  \includegraphics[width=\textwidth]{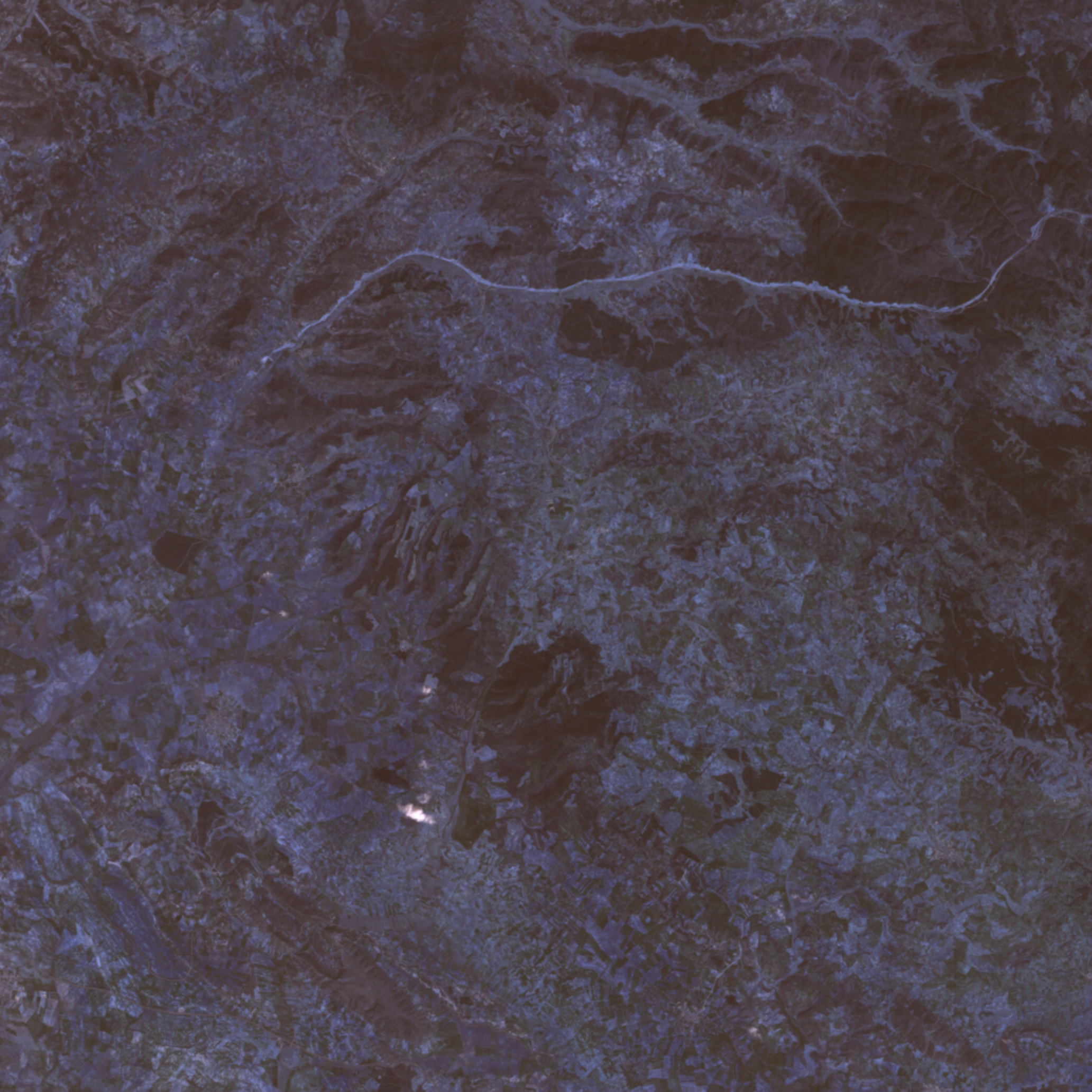}
  \caption{ConvLNP inpainting of 1024x1024 Kenya image using 64x64 patches.}
  \label{patches_convlnp}
\end{figure}

\begin{figure}[hbtp]
\centering
  \includegraphics[width=\textwidth]{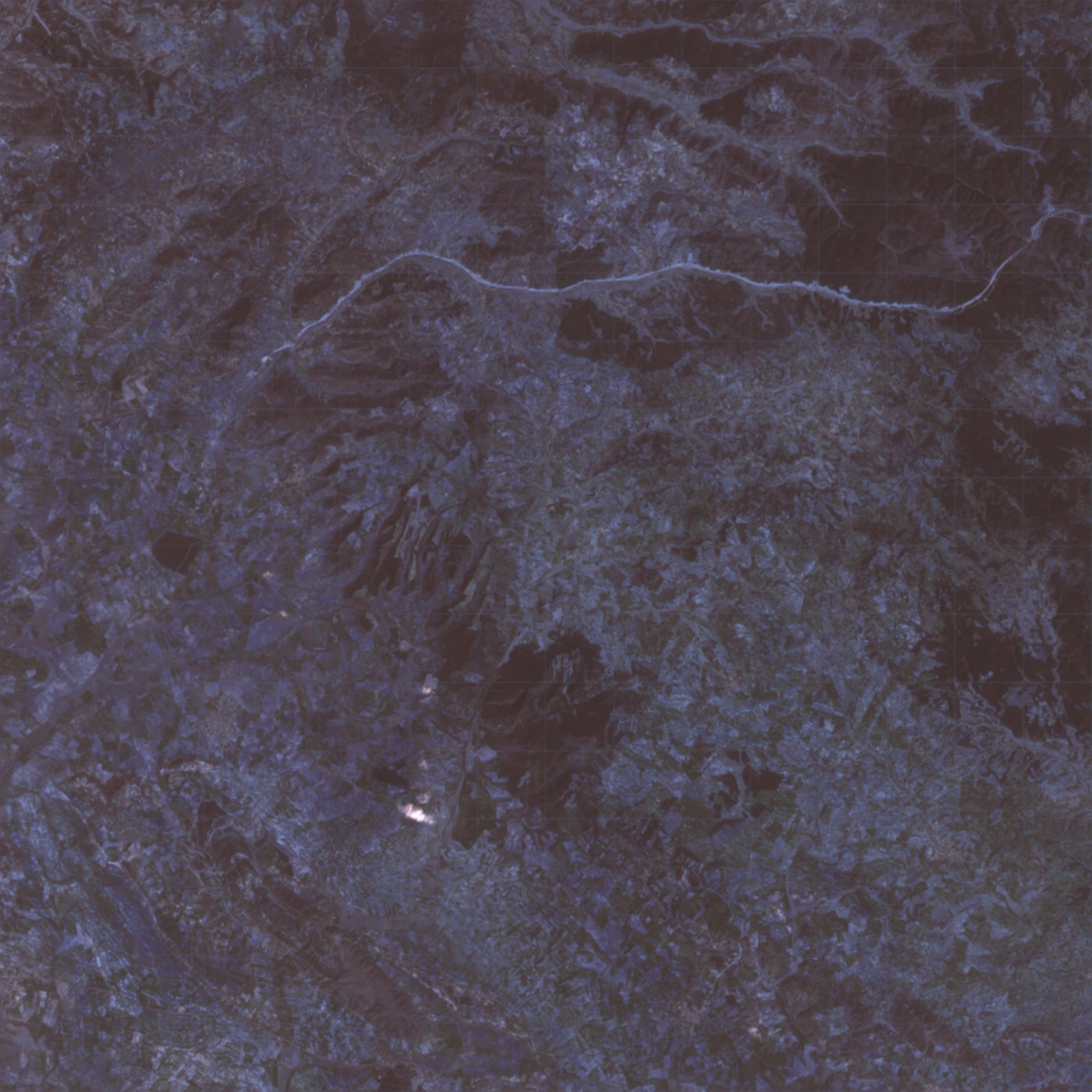}
  \caption{U-Net inpainting of 1024x1024 Kenya image using 64x64 patches.}
  \label{patches_unet}
\end{figure}

\begin{figure}[hbtp]
\centering
  \includegraphics[width=\textwidth]{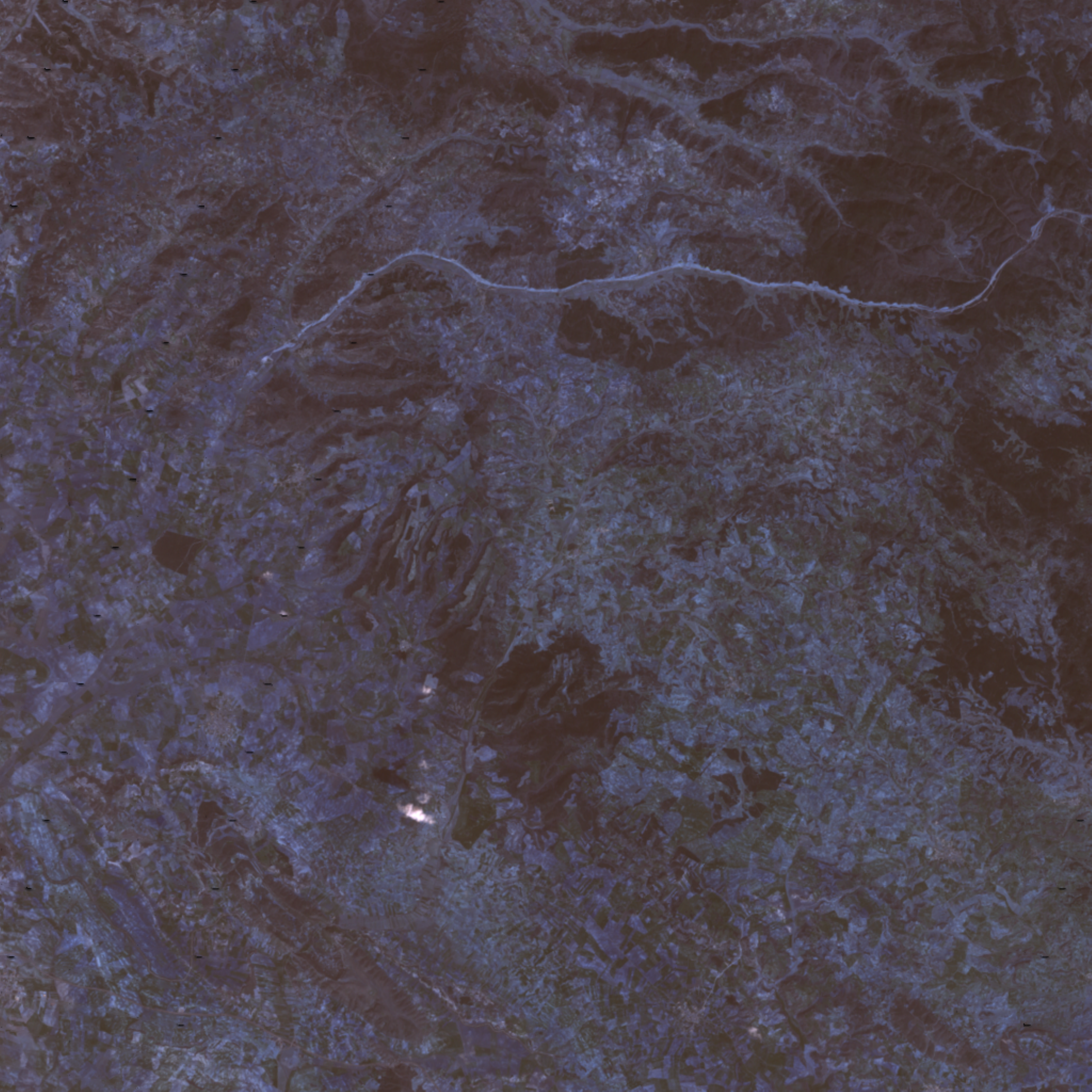}
  \caption{Navier-Stokes inpainting of 1024x1024 Kenya image using 64x64 patches.}
  \label{patches_ns}
\end{figure}

\begin{figure}[hbtp]
\centering
  \includegraphics[width=\textwidth]{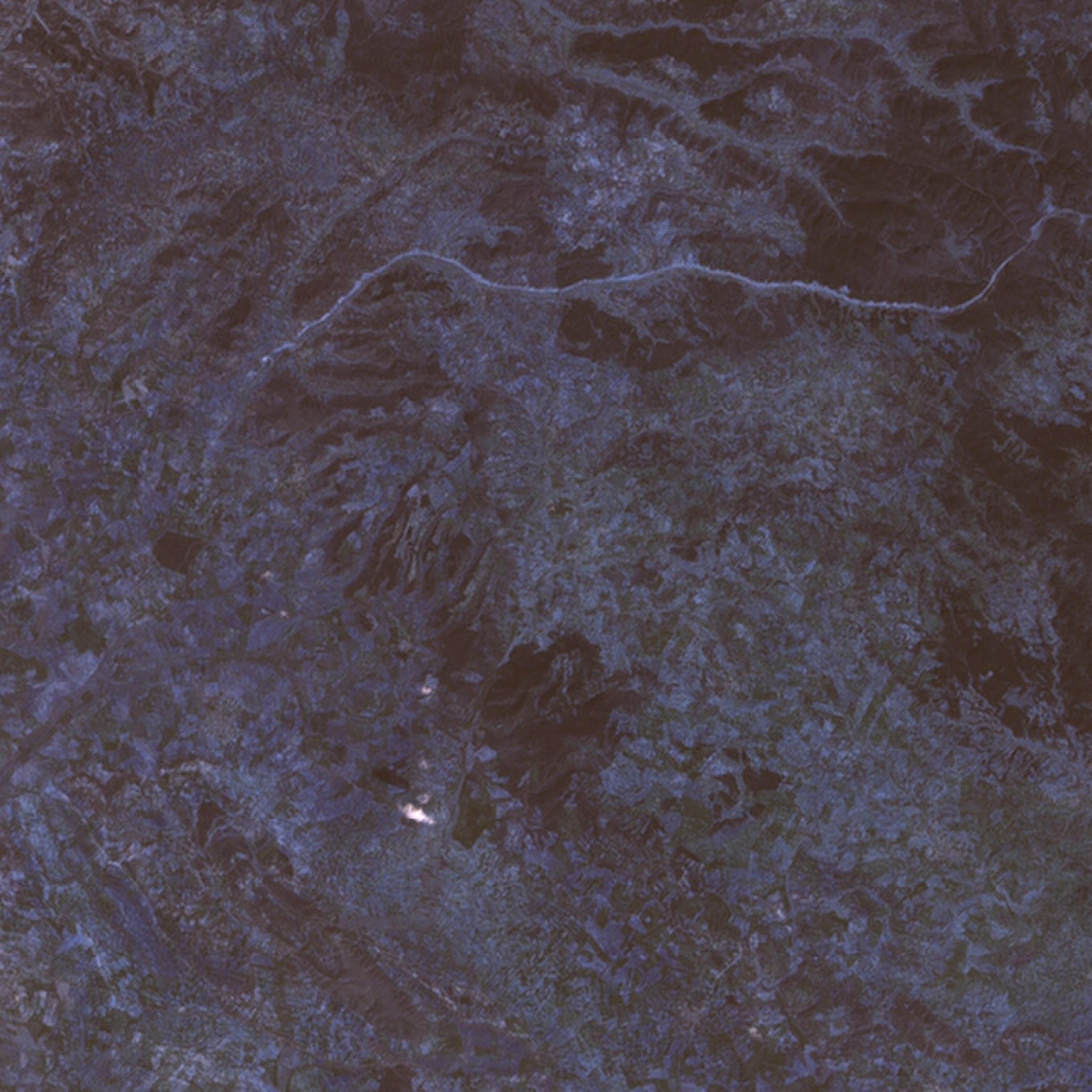}
  \caption{PartialConv inpainting of 1024x1024 Kenya image using 64x64 patches.}
  \label{patches_partialconv}
\end{figure}

\end{document}